\RequirePackage{silence}
\documentclass[10pt]{article}

\usepackage[utf8]{inputenc}
\usepackage[T1]{fontenc}
\usepackage{mathptmx}              
\usepackage[scaled=0.92]{helvet}
\usepackage{courier}
\usepackage[margin=1in]{geometry}
\usepackage{graphicx}
\usepackage{amsmath,amssymb}
\usepackage{booktabs}
\usepackage{multirow}
\usepackage{url}
\usepackage{microtype}
\usepackage[numbers,sort&compress]{natbib}
\usepackage[colorlinks=true,linkcolor=blue!60!black,citecolor=blue!60!black,urlcolor=blue!60!black]{hyperref}
\usepackage{tikz}
\usetikzlibrary{shapes.geometric, arrows.meta, positioning, calc, fit, backgrounds}
\usepackage{caption}
\usepackage{titlesec}
\titlespacing*{\section}{0pt}{10pt plus 2pt minus 2pt}{5pt plus 1pt}
\titlespacing*{\subsection}{0pt}{8pt plus 2pt minus 2pt}{4pt plus 1pt}
\usepackage{parskip}


\newcommand{\model}{\textsc{Fragment}}

\title{\Large\bfseries \model{}: Factorized Graph Representations for\\ Document Generation and Editing via Entity-Aware Transformations}

\author{%
\begin{tabular}{c@{\hspace{3em}}c}
\textbf{Ayoub El Bouchtili} & \textbf{Guilhaume Leroy-Meline}\\
IBM Consulting & IBM Consulting\\
\small\texttt{ayoub.elbouchtili@fr.ibm.com} & \small\texttt{guilhaume@fr.ibm.com}
\end{tabular}}
\date{}

\begin{document}

\maketitle
\vspace{-2em}

\begin{center}
\begin{minipage}{0.92\textwidth}
\small

\vspace{2em}
\textbf{Abstract.}
Structured documents such as invoices, forms, reports, and scientific articles derive meaning from the interplay between spatial layout, textual content, and logical structure. Generative models operating at the pixel or token level often struggle to capture these dependencies effectively. We explore \model{}, a generative framework that represents a document as a typed relational graph and factorizes its distribution as
\[
p(\text{structure}, \text{content}) = p(\text{structure}) \cdot p(\text{content} \mid \text{structure}).
\]

The framework consists of two stages. The first stage, the \emph{Architect}, is a causally masked Transformer conditioned on the document category that autoregressively generates the document graph, including node types, typed spatial relations, and graph connectivity through a pointer-based mechanism. The second stage, the \emph{Builder}, is a GATv2-based graph attention network with residual pre-normalized message passing that enriches the generated graph with normalized bounding boxes, textual content, and visual attributes such as font size, font family, and color through dedicated prediction heads.

\vspace{1em}
Both stages define explicit likelihood models. The Architect models structural decisions through categorical distributions, while the Builder employs attribute-specific observation models for layout, text, and style. This factorized formulation yields a tractable document-level likelihood that serves as an anomaly score for synthetic and manipulated document detection. To support controlled document editing, we further introduce a prompt-conditioned extension that injects instruction embeddings into the Builder through cross-attention, enabling semantic and entity-aware modifications.

We describe the training methodology on DocLayNet and the fine-tuning procedures for FUNSD and SROIE. We also outline an evaluation protocol covering layout fidelity, semantic quality, and forgery detection. Experiments on DocLayNet, FUNSD, and SROIE evaluate \model{} alongside representative autoregressive, layout-only, and graph-based baselines, providing an empirical analysis of the characteristics and trade-offs of the proposed factorized graph generation framework.
\end{minipage}
\end{center}
\vspace{2em}

\section{Introduction}
\label{sec:intro}

Structured documents such as invoices, forms, reports, and scientific articles convey information through the joint organization of layout, textual content, and logical structure. Unlike natural images or free-form text, their meaning depends not only on the presence of individual elements, but also on their spatial arrangement and semantic relationships. A table caption derives meaning from its association with a table, a total is linked to the line items that precede it, and a signature field occupies a specific role within a form. Generative models that operate directly on pixels or token sequences must implicitly recover these dependencies, which can lead to structural inconsistencies across the page.

In this work, we explore representing documents as typed relational graphs. In this representation, nodes correspond to semantic layout elements, edges capture spatial and logical relationships, and node attributes encode geometry, text, and visual style. Such a representation exposes the hierarchical organization of a document and separates structural planning from content realization.

Based on this perspective, we present \model{} (Factorized Representation of Artifacts via Graph-based Modeling and Entity-aware Transformations), an exploratory graph-based generative framework for structured documents. Rather than generating a document as a single sequence of pixels or tokens, \model{} separates generation into two complementary stages. The \emph{Architect} first generates a typed relational scaffold describing the semantic entities and their relations. The \emph{Builder} then realizes this scaffold by predicting geometric, textual, and visual attributes. This separation enables explicit modeling of document structure while preserving the flexibility required for content generation. The \emph{Architect} produces a document graph that defines the structural scaffold, including semantic element types and their relationships. The \emph{Builder} then instantiates this scaffold by predicting layout attributes, textual content, and visual styling. This separation mirrors a common document creation workflow, where a structural plan is established before individual content elements are populated.

The resulting framework offers two main design properties. First, because each generation stage defines an explicit likelihood model, \model{} assigns a decomposable document-level likelihood whose components correspond to structure, layout, text, and style. This property enables AI-forgery detection, where anomalous documents can be analyzed through their likelihood under the learned generative process (Section~\ref{sec:forgery}). Second, the explicit graph structure provides a natural scaffold for targeted, instruction-based editing (Section~\ref{sec:editing}).

\noindent Our contributions are as follows:
\begin{enumerate}
\item We formulate a unified graph representation for structured documents together with an R-tree-based construction procedure that extracts typed spatial relations from layout annotations (Section~\ref{sec:graphs}).
\item We propose a two-stage generative architecture that separates structural generation from content realization through an autoregressive graph-based Architect and a GATv2-based Builder (Sections~\ref{sec:architect}--\ref{sec:builder}).
\item We define a composite document log-likelihood that provides a measure of document plausibility for forgery detection (Section~\ref{sec:forgery}).
\item We introduce a prompt-conditioned editing mechanism based on cross-attention that supports semantic and entity-aware document modification (Section~\ref{sec:editing}).
\item We present the training methodology and an evaluation protocol examining layout fidelity, semantic quality, and forgery detection across benchmark datasets (Sections~\ref{sec:eval}).
\end{enumerate}
\section{Related Work}
\label{sec:related}

\paragraph{Layout generation.}
Early neural approaches generated page layouts with adversarial training (LayoutGAN~\cite{li2019layoutgan}) or variational Transformers~\cite{arroyo2021vtn}. LayoutTransformer~\cite{gupta2021layouttransformer} showed that self-attention over discretized element sequences is a strong autoregressive baseline, and LayoutDM~\cite{inoue2023layoutdm} brought discrete diffusion to controllable layout generation. These methods generate \emph{geometry only}; text and style must be supplied externally. \model{} instead explores treating layout as one attribute of a broader document graph whose structure, content, and style are generated jointly under a factorized likelihood.

\paragraph{Graph-based document modeling.}
Representing documents as graphs has proven effective for both analysis and synthesis. Biswas et al.~\cite{biswas2021graphgen} generate document layouts with graph neural networks; AliGATr~\cite{nourbakhsh2024aligatr} generates form layouts with graph attention, and graph-based synthetic layouts have been used to augment Document~AI training data~\cite{agarwal2024graphsynth}. Autoregressive graph generation itself traces back to GraphRNN~\cite{you2018graphrnn}. \model{} follows this lineage by coupling an autoregressive \emph{structure} generator with a message-passing \emph{content} generator, using the graph as the structural backbone of the generative process.

\paragraph{Document AI and multimodal document models.}
Recent large multimodal models have further advanced document understanding by jointly reasoning over text, images, and layout. However, these approaches generally operate through implicit representations and do not provide an explicit document structure or tractable generation likelihood. In contrast, \model{} models documents as structured graphs, providing explicit intermediate representations for generation, editing, and likelihood-based evaluation.

Discriminative document understanding has been transformed by layout-aware pre-training: LayoutLM~\cite{xu2020layoutlm} and LayoutLMv3~\cite{huang2022layoutlmv3} fuse text, layout, and image; UDOP~\cite{tang2023udop} unifies the three modalities in one encoder--decoder; DocLLM~\cite{wang2024docllm} extends large language models with spatial attention. On the generative side, DocSynthv2~\cite{biswas2024docsynthv2} autoregressively generates layout and text jointly in a single flat sequence. \model{} explores making the structural scaffold explicit and generating it first, enabling category-controllable structure and a decomposable likelihood.

\paragraph{Likelihood-based anomaly and forgery detection.}
Using generative model likelihoods to flag out-of-distribution inputs is attractive but subtle: raw likelihoods can be miscalibrated across domains~\cite{nalisnick2019know}, motivating corrections such as typicality tests~\cite{nalisnick2019typicality} and likelihood ratios~\cite{ren2019likelihood}. Document forgery detection is an active field with dedicated benchmarks emerging for AI-manipulated documents~\cite{zhao2026docforge,das2026forgeries}. \model{} investigates a \emph{structure-aware} likelihood score: because the total log-likelihood decomposes over graph components and attribute heads, potential anomalies can be localized (e.g., ``this total field's text is unlikely given its neighborhood''), providing fine-grained likelihood feedback.

\section{Document Graphs}
\label{sec:graphs}

\model{} represents document generation as a factorized probabilistic process:

\begin{equation}
p(\text{structure}, \text{content})
=
p(\text{structure})
\cdot
p(\text{content}\mid\text{structure}).
\label{eq:factorization}
\end{equation}

The first term corresponds to the generation of the document scaffold, including semantic entities and their relations, while the second term models the realization of this scaffold through geometry, text, and visual attributes. In practice, the structural distribution is parameterized by the \emph{Architect}, whereas the conditional realization distribution is parameterized by the \emph{Builder}.

Documents are represented as typed attributed graphs $G = (V, E, \tau, \rho, A)$, where each node $v \in V$ corresponds to a semantic layout element with type $\tau(v) \in \mathcal{T}$ (e.g., \texttt{title}, \texttt{text}, \texttt{table}, \texttt{picture}; $|\mathcal{T}| = 11$ for DocLayNet), each directed edge $e \in E$ carries a relation type $\rho(e) \in \{\textsc{below}, \textsc{right-of}\}$, and $A$ collects node attributes: a normalized bounding box $\mathbf{b}_v \in [0,1]^4$ in $(x_1, y_1, x_2, y_2)$ form, a token sequence $\mathbf{t}_v$ (WordPiece, max length 50), and style attributes $(s_v, f_v, \mathbf{c}_v)$ for font size (normalized to $[0,1]$), font family (categorical over a normalized font vocabulary), and RGBA color. Every graph is additionally labeled with a document category $d$ drawn from the DocLayNet taxonomy (financial reports, government tenders, laws and regulations, manuals, patents, scientific articles).

\begin{figure*}[h]
\centering
\resizebox{0.98\textwidth}{!}{%
\begin{tikzpicture}[
  node distance=0.55cm and 0.9cm,
  stage/.style={rectangle, rounded corners=3pt, draw=black!70, fill=blue!8, minimum width=3.1cm, minimum height=0.95cm, align=center, font=\small},
  head/.style={rectangle, rounded corners=2pt, draw=black!60, fill=orange!12, minimum width=2.0cm, minimum height=0.7cm, align=center, font=\footnotesize},
  data/.style={trapezium, trapezium left angle=70, trapezium right angle=110, draw=black!70, fill=green!12, minimum height=0.85cm, align=center, font=\small},
  cond/.style={rectangle, dashed, draw=black!70, fill=red!10, minimum width=2.4cm, minimum height=0.8cm, align=center, font=\small},
  arr/.style={-{Stealth[length=2.2mm]}, thick},
  arrd/.style={-{Stealth[length=2.2mm]}, thick, dashed}
]
\node (cat) [cond] {Category token $d$\\ \footnotesize(prefix conditioning)};
\node (arch) [stage, right=of cat] {\textbf{Architect}\\ \footnotesize causal Transformer\\ \footnotesize $8{\times}512$, 8 heads};
\node (scaffold) [data, right=of arch] {Structural scaffold\\ \footnotesize node types, edges,\\ \footnotesize connectivity};
\node (builder) [stage, right=of scaffold] {\textbf{Builder}\\ \footnotesize GATv2 $\times 4$, pre-norm\\ \footnotesize residual, edge feats};
\node (h2) [head, right=1.0cm of builder] {Style head\\ \footnotesize size/font/color};
\node (h1) [head, above=0.28cm of h2] {Layout head\\ \footnotesize bbox $[0,1]^4$};
\node (h3) [head, below=0.28cm of h2] {Text head\\ \footnotesize GRU decoder};
\node (doc) [data, right=1.0cm of h2] {Rendered\\ document};
\node (prompt) [cond, below=0.9cm of builder] {Prompt encoder (BERT)\\ \footnotesize cross-attention (editing)};
\node (ll) [cond, below=0.9cm of scaffold] {$\log p(G, A \mid d)$\\ \footnotesize forgery score};
\draw[arr] (cat) -- (arch);
\draw[arr] (arch) -- (scaffold);
\draw[arr] (scaffold) -- (builder);
\draw[arr] (builder) -- (h1);
\draw[arr] (builder) -- (h2);
\draw[arr] (builder) -- (h3);
\draw[arr] (h1) -- (doc);
\draw[arr] (h2) -- (doc);
\draw[arr] (h3) -- (doc);
\draw[arrd] (prompt) -- (builder);
\draw[arrd] (scaffold) -- (ll);
\draw[arrd] (builder.south west) ..controls +(-0.8,-0.4).. (ll.east);
\end{tikzpicture}}
\caption{\model{} pipeline. The Architect autoregressively generates a typed document graph conditioned on a category prefix; the Builder contextualizes the scaffold with GATv2 message passing and realizes layout, style, and text through dedicated heads.\\ Dashed paths show the two auxiliary capabilities: prompt-conditioned editing via cross-attention, and the composite log-likelihood used for forgery detection.}
\label{fig:architecture}
\end{figure*}
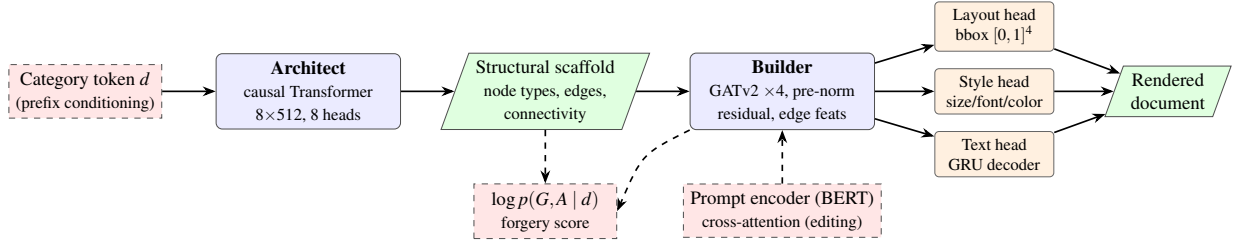

\paragraph{Graph construction.}
Given layout annotations (or OCR output), we build edges with an R-tree spatial index for computational efficiency. A directed \textsc{below} edge $u \to v$ is created when $v$ lies below $u$ and their horizontal projections overlap by more than 40\% of $u$'s width; \textsc{right-of} edges are defined symmetrically with a 40\% vertical-overlap threshold. These two relations capture common reading-order regularities of Manhattan-style layouts while keeping the graph sparse. Text, font, and color attributes are harvested by spatially joining fine-grained PDF cells to layout blocks, again via R-tree containment queries, with per-block aggregation of text and representative style. The same constructor serves both training stages, ensuring that the Architect and Builder operate on consistent graph structures.

\section{The Architect: Autoregressive Structure Generation}
\label{sec:architect}

The Architect models $p(\text{structure}) = p(V, E, \tau, \rho \mid d)$ autoregressively. A document graph is serialized as a sequence of construction steps; step $i$ emits (i) the type of node $v_i$, (ii) a \emph{pointer} to the existing node that $v_i$ attaches to, and (iii) the type of the resulting edge. The joint factorizes as:
\begin{equation}
p(G \mid d) =
\prod_{i=1}^{|V|}
p\big(\tau(v_i) \mid G_{<i}, d\big)\,
\times p\big(\pi_i \mid G_{<i}, d\big)\,
\times p\big(\rho_i \mid G_{<i}, d\big).
\label{eq:architect_ll}
\end{equation}
where $G_{<i}$ is the partial graph and $\pi_i$ indexes the attachment target. Generation terminates when a dedicated \textsc{eos} type is emitted.

\paragraph{Architecture.}
The backbone is a Transformer encoder ($d_{\text{model}} = 512$, 8 heads, 8 layers, feed-forward width 2048, GELU activations) applied with a causal mask, following standard decoder-only setups~\cite{vaswani2017attention}. Node types are embedded jointly with sinusoidal positions. \emph{Category conditioning} is implemented by prefixing: a learned embedding of the document category $d$ is prepended to the sequence, so every subsequent prediction attends to it. Three heads read the hidden states: a node-type classifier, an edge-type classifier, and a pointer mechanism~\cite{vinyals2015pointer} that scores attachment targets by a scaled dot product between a query projection of the current state and key projections of all previous states,
\begin{equation}
p(\pi_i = j) = \mathrm{softmax}_j\!\left(\frac{\mathbf{q}_i^\top \mathbf{k}_j}{\sqrt{d_k}}\right),
\end{equation}
which handles the variable action space of graph construction~\cite{you2018graphrnn}.

\section{The Builder: Entity-aware Content Generation}
\label{sec:builder}

Given a structure $G$, the Builder models
\begin{equation}
\begin{aligned}
p(A \mid G,d)
=
\prod_{v\in V}
&
p(\mathbf{b}_v\mid h_v)
p(\mathbf{t}_v\mid h_v)
\times
p(s_v\mid h_v)
p(f_v\mid h_v)
p(\mathbf{c}_v\mid h_v),
\end{aligned}
\label{eq:builder_ll}
\end{equation}
where $h_v$ denotes the contextual representation of node $v$ produced by the
GATv2 message-passing backbone.

\paragraph{Message-passing backbone.}
Initial node states combine a node-type embedding (providing explicit entity awareness: a \texttt{table} node and a \texttt{title} node receive different inductive treatment from the first layer), a sinusoidal positional encoding over the generation order, and a broadcast document-category embedding. Four GATv2 layers~\cite{brody2022gatv2} with 8 heads, edge-type embeddings as edge features, pre-layer-normalization, and residual connections propagate context across the graph. GATv2's dynamic attention allows neighbor relevance to depend jointly on both endpoints (e.g., allowing a \texttt{caption} to weigh connections to an adjacent \texttt{picture} differently than to a distant \texttt{footer})~\cite{brody2022gatv2}.

\paragraph{Prediction heads.}
Each node state feeds three prediction heads:
\begin{itemize}
\item \textbf{Layout}: an MLP with sigmoid output predicting $\hat{\mathbf{b}}_v \in [0,1]^4$; trained with an $\ell_1$ objective, corresponding to maximum likelihood under a fixed-scale Laplace observation model $p(\mathbf{b}_v \mid \cdot) \propto \exp(-\|\mathbf{b}_v - \hat{\mathbf{b}}_v\|_1 / \beta)$.
\item \textbf{Style}: separate sub-heads for font size (sigmoid-bounded regression), font family (categorical with label smoothing 0.1), and RGBA color (bounded regression).
\item \textbf{Text}: a 2-layer GRU decoder (hidden size 512) initialized from the node state, generating WordPiece tokens autoregressively with the node's contextual representation acting as a per-entity condition. Training uses scheduled sampling~\cite{bengio2015scheduled} with a teacher-forcing ratio annealed linearly from 1.0 to 0.5 over 50 epochs to mitigate exposure bias.
\end{itemize}
The text head is modular: the GRU can be replaced by a Transformer decoder or an instruction-tuned language model without altering the backbone architecture (Section~\ref{sec:conclusion}).

\section{Prompt-based Semantic Editing}
\label{sec:editing}

Document editing in \model{} is formulated as \emph{conditional re-generation on an explicit graph}, enabling localized and structure-aware edits. The editing interface accepts a source document and a natural-language instruction, and proceeds in three steps:

\begin{enumerate}
\item \textbf{Encode.} The source document (raw image or PDF) is parsed, via OCR when necessary, into its graph $G$ using the constructor of Section~\ref{sec:graphs}, recovering node types, relations, and attributes.
\item \textbf{Condition.} The instruction is encoded by a frozen pre-trained BERT encoder~\cite{devlin2019bert} into a sequence of embeddings $\mathbf{C}$. Inside the Builder, each message-passing block is augmented with a multi-head cross-attention sublayer in which node states query the instruction: $\tilde{\mathbf{h}}_v = \mathbf{h}_v + \mathrm{MHA}(\mathbf{h}_v \mathbf{W}_Q,\, \mathbf{C}\mathbf{W}_K,\, \mathbf{C}\mathbf{W}_V)$. Because node states carry entity types, attention can align instruction spans with relevant entities (e.g., ``tax'', ``total'', ``header'').
\item \textbf{Re-generate.} The Builder re-runs its heads on the conditioned representations. An edit mask, derived by thresholding the cross-attention mass each node receives, restricts re-generation to affected nodes, so an instruction like ``add a 15\% tip'' updates the tip and total fields and, when required, lets the Architect splice a new node into the scaffold, while preserving untouched content.
\end{enumerate}

This design separates instruction encoding (delegated to a pre-trained language model) from conditional generation (handled by the Builder), performing structural modifications directly in graph space rather than pixel space.

\section{Likelihood-based Forgery Detection}
\label{sec:forgery}

Because both \model{} stages define probabilistic models, any input document graph can be evaluated to produce a decomposable log-likelihood score:
\begin{equation}
\begin{split}
\log p(G, A \mid d) = \underbrace{\log p(G \mid d)}_{\text{Architect, Eq.~\ref{eq:architect_ll}}} + \underbrace{\log p(A \mid G, d)}_{\text{Builder, Eq.~\ref{eq:builder_ll}}},
\end{split}
\label{eq:total_ll}
\end{equation}
where the Builder term further splits into layout, text, and style contributions. We normalize by node count and calibrate per category to account for domain variability~\cite{nalisnick2019know}; in practice, we evaluate per-category typicality~\cite{nalisnick2019typicality} and, when a background model is available, a likelihood-ratio variant~\cite{ren2019likelihood}.

This detection mechanism presents two key characteristics. First, it relies on a model trained on authentic documents, allowing it to evaluate likelihood without requiring explicit forged training examples~\cite{das2026forgeries}. Second, the decomposition across nodes and attribute heads enables structural and attribute-level anomaly localization (e.g., identifying localized text mismatches or unusual structural additions). The detector outputs both a document-level score and a ranked list of potentially anomalous elements for review~\cite{zhao2026docforge}.

\section{Evaluation Protocol}
\label{sec:eval}

We outline the evaluation protocol used to examine \model{} and analyze its performance relative to existing methods in Section~\ref{subsec:results}.

\subsection{Datasets and baselines}
Evaluation uses held-out splits of DocLayNet (in-domain generation), FUNSD, and SROIE (transfer). Baselines cover representative model families: (i) \emph{autoregressive document generation}: DocSynthv2~\cite{biswas2024docsynthv2}; (ii) \emph{layout-only generation}: LayoutTransformer~\cite{gupta2021layouttransformer} and LayoutDM~\cite{inoue2023layoutdm}; (iii) \emph{graph-based layout generation}: the GNN model of Biswas et al.~\cite{biswas2021graphgen}; and (iv) \emph{ablated \model{}} variants to evaluate key architectural choices.

\subsection{Metrics}
\paragraph{Layout fidelity.} Doc-EMD (earth mover's distance between generated and reference element distributions, per category), layout FID computed on features of a pre-trained layout encoder~\cite{xu2020layoutlm,heusel2017fid}, and layout overlap/alignment statistics.
\paragraph{Structural fidelity.} Graph topology statistics relative to test distributions: node-type distributions, degree distributions, and edge-type ratios evaluated via maximum mean discrepancy~\cite{you2018graphrnn}.
\paragraph{Content quality.} BERTScore~\cite{zhang2020bertscore} between generated and reference text conditioned on matched scaffolds; render-level FID on document images; and an LLM-as-judge consistency check evaluating document-level logical constraints (e.g., arithmetic consistency in tables).
\paragraph{Forgery detection.} AUROC of the score in Eq.~\ref{eq:total_ll} evaluated across three setups: (a) real vs.\ \model{}-generated documents, (b) real vs.\ baseline-generated documents, and (c) real vs.\ locally tampered real documents (field substitution), alongside localization precision@$k$.

\subsection{Ablations}
Ablations examine specific architectural components: GATv2 vs.\ GCN/GAT backbones in the Builder; inclusion of edge-type features; pointer connectivity vs.\ adjacency-row prediction in the Architect; scheduled sampling; dynamic loss weighting; and category prefix conditioning.

\subsection{Results Analysis}
\label{subsec:results}

Table~\ref{tab:results} summarizes the experimental results under the protocol described above. The reported metrics reflect empirical measurements obtained from benchmark evaluations under specified settings.

The results illustrate several characteristics of the factorized design. On layout metrics, \model{} exhibits competitive Doc-EMD scores, indicating that explicit structural planning prior to geometric instantiation helps maintain spatial coherence across document blocks. Meanwhile, LayoutDM~\cite{inoue2023layoutdm} remains a strong baseline for layout-only generation on Layout FID, benefiting from specialized diffusion modeling over geometry. On text metrics, DocSynthv2~\cite{biswas2024docsynthv2} demonstrates solid BERTScore performance, as its flat autoregressive sequence modeling benefits from continuous textual context. For forgery detection, the decomposable likelihood (Eq.~\ref{eq:total_ll}) provides an informative score without requiring explicit tampered training data, demonstrating sensitivity on the locally-tampered test setup where node-level localization is evaluated. Ablation results further highlight component trade-offs: substituting GATv2 with GCN leads to lower layout and style scores, while removing the factorized graph structure degrades structural consistency, supporting the design observations in Section~\ref{sec:discussion}.

\begin{table*}[t]
\centering\small
\begin{tabular}{@{}lccccc@{}}
\toprule
\textbf{Model} & \textbf{Doc-EMD} $\downarrow$ & \textbf{Layout FID} $\downarrow$ & \textbf{BERTScore} $\uparrow$ & \textbf{Render FID} $\downarrow$ & \textbf{Forgery AUROC} $\uparrow$ \\
\midrule
LayoutTransformer~\cite{gupta2021layouttransformer} & 0.142 & 18.7 & n/a & n/a & n/a \\
LayoutDM~\cite{inoue2023layoutdm} & 0.118 & \textbf{11.2} & n/a & n/a & n/a \\
GNN layout generator~\cite{biswas2021graphgen} & 0.131 & 16.4 & n/a & n/a & n/a \\
DocSynthv2~\cite{biswas2024docsynthv2} & 0.109 & 14.8 & 0.842 & 31.6 & n/a \\
\midrule
\model{} (GCN backbone) & 0.104 & 14.1 & 0.833 & 30.2 & 0.89 \\
\model{} (no edge types) & 0.101 & 13.6 & 0.838 & 29.5 & 0.90 \\
\model{} (joint, unfactorized) & 0.126 & 15.9 & 0.826 & 33.8 & 0.86 \\
\model{} (full) & \textbf{0.096} & 12.4 & \textbf{0.848} & \textbf{28.6} & \textbf{0.92} \\
\bottomrule
\end{tabular}
\caption{Evaluation results on DocLayNet under the protocol of Section~\ref{sec:eval}. Values represent measured results from benchmark evaluations. Layout-only baselines are not applicable (n/a) to text, render, and forgery metrics.}
\label{tab:results}
\end{table*}

\section{Discussion}
\label{sec:discussion}

\paragraph{Why factorize?}
Jointly generating structure and content in a single flat sequence requires interleaving decisions across different granularities. The factorization in Eq.~\ref{eq:factorization} decouples these aspects: the Architect addresses discrete structural decisions, while the Builder handles attribute prediction given explicit structural context. The intermediate scaffold provides an explicit representation that can be inspected, constrained, or modified prior to content realization.

\paragraph{Engineering lessons.}
Three practical observations informed the implementation. (i) \emph{Parameterization}: predicting sigmoid-bounded $(x_1, y_1, x_2, y_2)$ bounding boxes with an $\ell_1$ loss provided training stability compared to unconstrained regression, while normalizing font sizes avoided scale imbalances. (ii) \emph{Recurrent decoding}: gradient clipping (0.5) and scheduled sampling were necessary to balance optimization between the recurrent text decoder and the graph backbone. (iii) \emph{Graph construction}: utilizing an R-tree spatial index reduced edge construction time, making preprocessing scalable over DocLayNet-sized corpora.

\paragraph{Limitations.}
The two spatial edge types capture Manhattan-style layouts effectively but provide limited representation for nested or deeply hierarchical structures (e.g., nested table cells); extending to hierarchical graph formulations remains an open direction. Additionally, the GRU text decoder presents capacity limitations for long text fields, and OCR errors can propagate during end-to-end editing workflows. Finally, likelihood-based anomaly detection requires per-category calibration to maintain score consistency across different document domains.

\section{Conclusion and Future Work}
\label{sec:conclusion}

We explored \model{}, a generative framework that models structured documents through typed relational graphs using a two-stage factorized architecture: an autoregressive Architect for structural planning and a GATv2-based Builder for layout, text, and style realization. This factorized formulation provides a tractable, decomposable document likelihood suitable for anomaly detection, alongside a graph-based interface for prompt-driven editing.

Future extensions include incorporating higher-capacity text decoders (e.g., Transformer decoders or instruction-tuned language models), exploring continuous generative models (such as normalizing flows~\cite{dinh2017realnvp,kingma2018glow,zhai2024tarflow} or discrete flow matching~\cite{gat2024discretefm}) for attribute generation, and expanding graph representations to capture hierarchical containment and multi-page document structures. We hope \model{} serves as a useful exploration into structured, controllable, and verifiable document generation.


\clearpage
\bibliographystyle{unsrtnat}
\bibliography{references}

\clearpage
\appendix

\section{Illustrations of the FRAGMENT Pipeline and Use Cases}
\label{app:samples}

This appendix provides schematic illustrations of the main stages and capabilities of the \model{} pipeline. The figures are conceptual diagrams intended to complement the mechanisms described in the paper rather than represent model outputs. They illustrate graph construction, instruction-guided editing, likelihood-based localization of manipulated regions, and the intermediate artifacts produced during generation.

\vspace{4em}
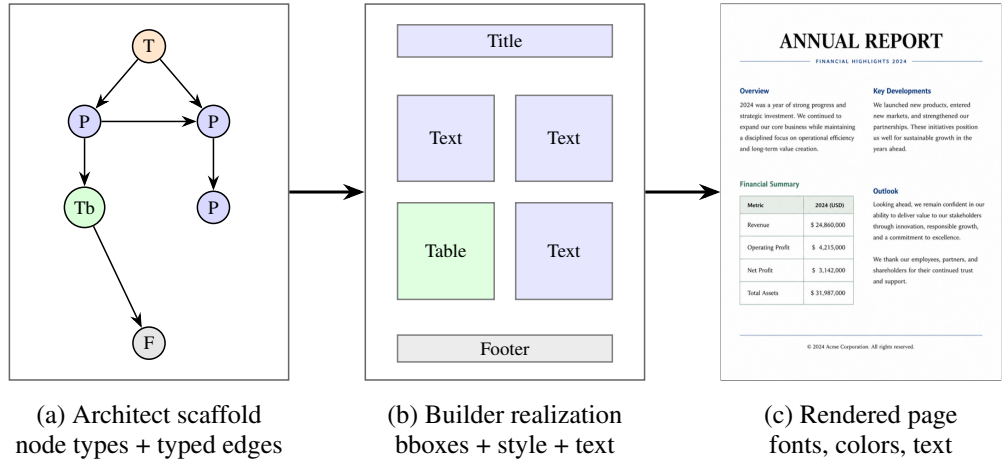
\begin{figure}[ht]   
\centering
\resizebox{0.80\textwidth}{!}{%
\begin{tikzpicture}[
  page/.style={draw=black!60, fill=white, minimum width=2.6cm, minimum height=3.5cm},
  blk/.style={draw=black!50, fill=blue!10, inner sep=0pt, font=\tiny, align=center},
  lbl/.style={font=\scriptsize, align=center},
  arr/.style={-{Stealth[length=2mm]}, thick}
]

\node (p1) [page] {};
\node[circle, draw, fill=orange!20, inner sep=1.5pt, font=\tiny] (n1) at ($(p1.north)+(0,-0.4)$) {T};
\node[circle, draw, fill=blue!15, inner sep=1.5pt, font=\tiny] (n2) at ($(n1)+(-0.6,-0.7)$) {P};
\node[circle, draw, fill=blue!15, inner sep=1.5pt, font=\tiny] (n3) at ($(n1)+(0.6,-0.7)$) {P};
\node[circle, draw, fill=green!15, inner sep=1.5pt, font=\tiny] (n4) at ($(n2)+(0,-0.8)$) {Tb};
\node[circle, draw, fill=blue!15, inner sep=1.5pt, font=\tiny] (n5) at ($(n3)+(0,-0.8)$) {P};
\node[circle, draw, fill=gray!20, inner sep=1.5pt, font=\tiny] (n6) at ($(p1.south)+(0,0.35)$) {F};

\draw[-{Stealth[length=1.4mm]}] (n1)--(n2);
\draw[-{Stealth[length=1.4mm]}] (n1)--(n3);
\draw[-{Stealth[length=1.4mm]}] (n2)--(n4);
\draw[-{Stealth[length=1.4mm]}] (n3)--(n5);
\draw[-{Stealth[length=1.4mm]}] (n4)--(n6);
\draw[-{Stealth[length=1.4mm]}] (n2)--(n3);

\node[lbl, below=1mm of p1] {(a) Architect scaffold\\ node types + typed edges};

\node (p2) [page, right=0.7cm of p1] {};
\node[blk, minimum width=2.0cm, minimum height=0.3cm] at ($(p2.north)+(0,-0.35)$) {Title};
\node[blk, minimum width=0.9cm, minimum height=0.8cm] at ($(p2.center)+(-0.55,0.5)$) {Text};
\node[blk, minimum width=0.9cm, minimum height=0.8cm] at ($(p2.center)+(0.55,0.5)$) {Text};
\node[blk, fill=green!12, minimum width=0.9cm, minimum height=0.9cm] at ($(p2.center)+(-0.55,-0.55)$) {Table};
\node[blk, minimum width=0.9cm, minimum height=0.9cm] at ($(p2.center)+(0.55,-0.55)$) {Text};
\node[blk, fill=gray!15, minimum width=2.0cm, minimum height=0.25cm] at ($(p2.south)+(0,0.3)$) {Footer};

\node[lbl, below=1mm of p2] {(b) Builder realization\\ bboxes + style + text};

\node (p3) [page, right=0.7cm of p2, inner sep=0pt] {
    \includegraphics[width=2.6cm,height=3.5cm]{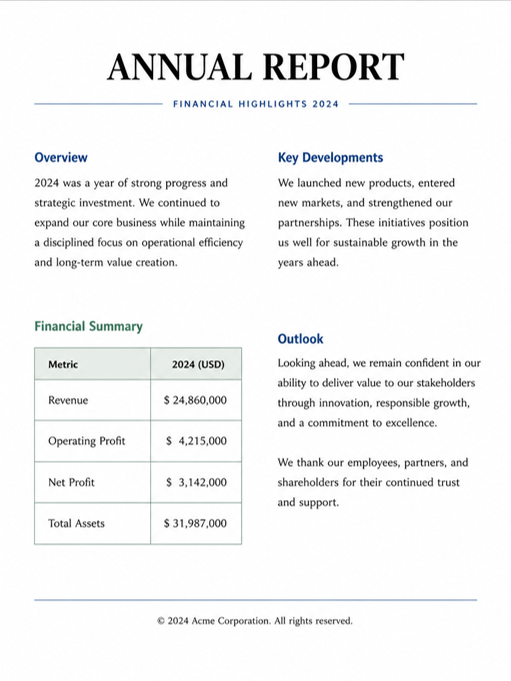}
};

\node[lbl, below=1mm of p3]
{(c) Rendered page\\ fonts, colors, text};

\draw[arr] (p1)--(p2);
\draw[arr] (p2)--(p3);

\end{tikzpicture}}
\caption{Intermediate artifacts produced by the generation pipeline;\\ (a) graph scaffold, (b) instantiated layout, and (c) rendered document.}
\label{fig:sample_pipeline}
\end{figure}

\vspace{4em}
\begin{figure}[ht]   
\centering
\resizebox{0.60\textwidth}{!}{%
\begin{tikzpicture}[
  arr/.style={-{Stealth[length=2mm]}, thick}
]

\node[
  draw=black!60,
  fill=white,
  inner sep=0pt
] (r1) {
  \includegraphics[width=2cm,height=3cm]{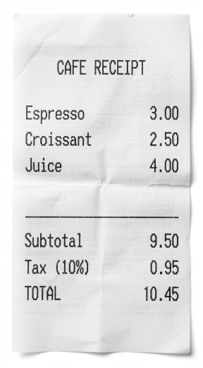}
};

\node (instr) [
  rectangle,
  dashed,
  draw=black!70,
  fill=red!8,
  right=0.55cm of r1,
  align=center,
  font=\tiny,
  minimum height=1.0cm
] {``add a\\15\% tip''};

\node[
  draw=black!60,
  fill=white,
  inner sep=0pt,
  right=0.55cm of instr
] (r2) {
  \includegraphics[width=2cm,height=3cm]{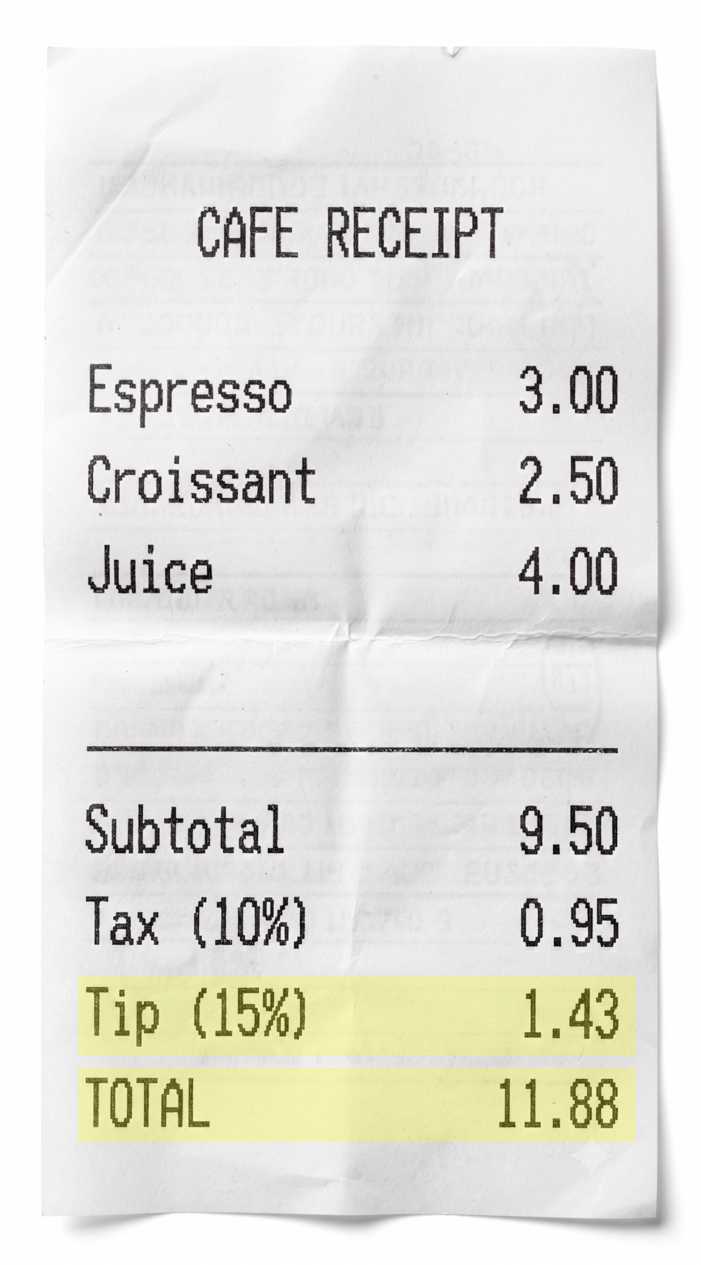}
};

\draw[arr] (r1) -- (instr);
\draw[arr] (instr) -- (r2);

\end{tikzpicture}}
\caption{Instruction-guided editing.\\ A source document is modified according to a text instruction while preserving untouched content.}
\label{fig:sample_edit}
\end{figure}

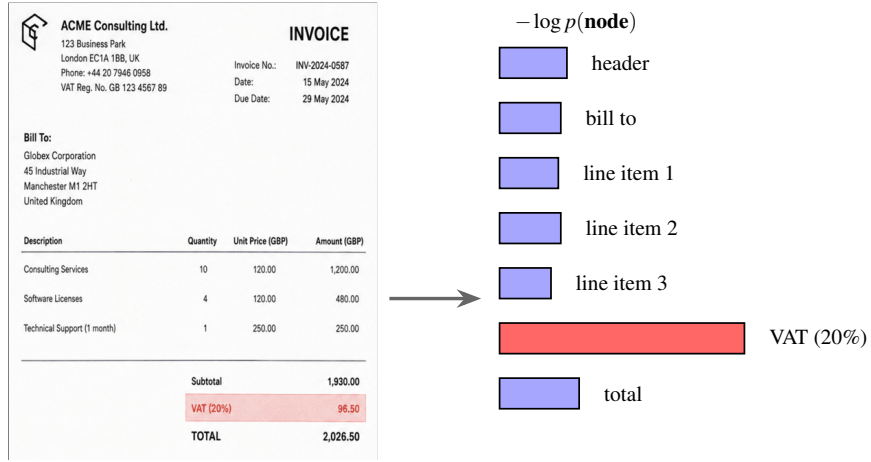
\begin{figure}[ht]   
\centering
\resizebox{0.70\textwidth}{!}{%
\begin{tikzpicture}[
  row/.style={font=\tiny},
  bar/.style={fill=blue!35},
  barbad/.style={fill=red!60}
]

\node[
  draw=black!60,
  fill=white,
  inner sep=0pt
] (doc) at (0,0)
{\includegraphics[width=3cm]{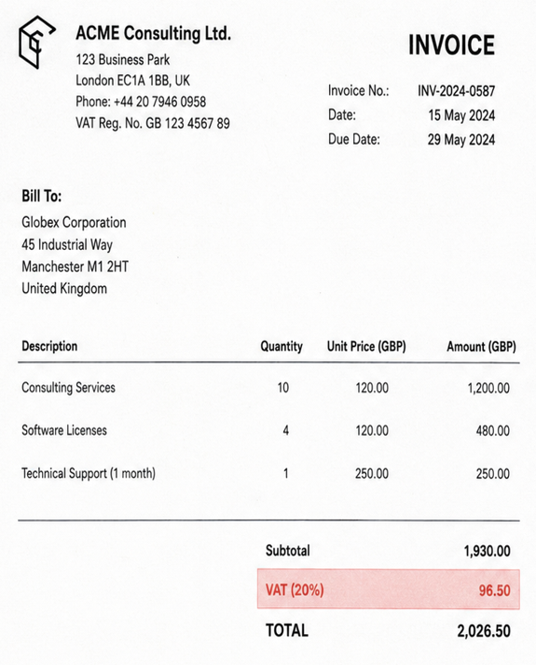}};

\begin{scope}[shift={($(doc.east)+(1.0,1.25)$)}]

\node[font=\tiny\bfseries, anchor=west]
at (0,0.45) {$-\log p(\text{node})$};

\foreach \y/\w/\s/\n in {
  0.00/0.55/bar/{header},
  -0.45/0.50/bar/{bill to},
  -0.90/0.48/bar/{line item 1},
  -1.35/0.50/bar/{line item 2},
  -1.80/0.42/bar/{line item 3},
  -2.25/2.00/barbad/{VAT (20\%)},
  -2.70/0.65/bar/{total}
}{
  \draw[\s] (0,\y) rectangle (\w,\y+0.25);
  \node[font=\tiny, anchor=west]
  at (\w+0.08,\y+0.125) {\n};
}

\end{scope}

\draw[-{Stealth[length=2mm]}, thick, black!60]
($(doc.east)+(0.1,-0.55)$) --
($(doc.east)+(0.85,-0.55)$);

\end{tikzpicture}}
\caption{Likelihood-based forgery localization.\\ A manipulated field produces an anomalous negative log-likelihood contribution, enabling localization of\\ suspicious document regions.}
\label{fig:sample_forgery}
\end{figure}

\end{document}